\pdfoutput=1
\documentclass[runningheads]{llncs}

\usepackage[utf8]{inputenc}
\usepackage[T1]{fontenc}
\usepackage{graphicx}
\usepackage{amsmath,amssymb}
\usepackage{booktabs}
\usepackage{placeins}
\usepackage{url}
\usepackage[hidelinks]{hyperref}
\hypersetup{
  pdftitle={AGO AI Quality Gate: Evidence-First Release Decisions for
    Retrieval-Augmented Generation},
  pdfauthor={Giulio Zeloni, Enrico Lo Conte, Salvatore Rionero,
    Giuseppe Santoro, Alessandro Rastelli, Fabio Sorrentino},
}

\newcommand{\sysnamefull}{AGO AI Quality Gate}
\newcommand{\sysname}{AGO}
\newcommand{\spromote}{\textsf{promote}}
\newcommand{\sreview}{\textsf{manual review}}
\newcommand{\sblock}{\textsf{block}}
\newcommand{\snoteval}{\textsf{not evaluable}}
\newcommand{\ci}[2]{{\footnotesize[#1, #2]}}
\begin{document}

\title{\sysnamefull{}: Evidence-First Release Decisions for
Retrieval-Augmented Generation}
\titlerunning{\sysnamefull{}}

\author{Giulio Zeloni \and Enrico Lo Conte \and \\
Salvatore Rionero \and Giuseppe Santoro \and \\ 
Alessandro Rastelli   \and Fabio Sorrentino}
\authorrunning{G. Zeloni et al.}
\institute{Protom Group S.p.A., Napoli, Italy\\
\email{giulio.zeloni@protom.com}}

\maketitle

\begin{abstract}
Enterprises adopting retrieval-augmented generation (RAG) face a
recurring operational decision: promote, revise, or block a system
version. The evidence is incomplete and the metrics come from fallible
LLM judges. We report on \sysnamefull{} (\sysname{}), an
evidence-first quality-gate framework deployed in industrial RAG
assessment engagements. \sysname{} integrates four key components: a four-state decision model that treats missing data and judge errors as explicit outcomes; layered scoring combining deterministic checks, local guardrails, and structured LLM evaluation; a stratified beta-binomial gate that quantifies regression risk probabilistically; and a mandatory meta-evaluation protocol to validate the LLM judge before it influences decisions. Since engagement data is proprietary, we
evaluate the judge layer on RAGBench, a public benchmark of 100k
annotated RAG traces across 12 datasets. On identical stratified test
samples ($N{=}1200$ per judge), a low-cost judge (gpt-4.1-nano) detects
non-adherent answers barely above chance (AUROC $0.603$
\ci{0.570}{0.634}), despite producing flawless protocol output, while
gpt-4o reaches $0.783$ \ci{0.756}{0.807}---yet its per-domain
performance still ranges from $0.62$ to $0.88$. A fixed-seed gate study
spanning regression, no change, and improvement quantifies unsafe
promotion, false-alarm cost, and improvement throughput. Under
regression, the decision-grade profile reduces unsafe promotion to
$22.2\%$--$35.1\%$, against $29.3\%$--$41.8\%$ for a naive gate. These
results support the design choices that judge quality must be measured
per engagement and that point estimates alone are not a release
decision.

\end{abstract}

\section{Introduction}\label{sec:intro}

Retrieval-augmented generation (RAG)~\cite{lewis2020retrieval} is now a
standard architecture for grounding large language models (LLMs) in
private document collections; deployments range from customer support
and compliance to healthcare and finance~\cite{gao2023survey}. Grounding
mitigates but does not eliminate hallucination: answers may contradict
retrieved evidence, cite the wrong source, or fabricate content when
retrieval fails~\cite{ji2023survey}. In an enterprise setting this
becomes a concrete, recurring decision about whether a given version can be released, or it needs additional reviews.

This paper reports on \sysnamefull{} (\sysname{}), a self-hosted
quality-gate platform built to support that decision. \sysname{} is
deployed at an IT consulting firm and used in RAG assessment
engagements with enterprise clients. Engagement constraints shaped the
design: client data cannot leave the premises; observability is often
incomplete; evaluation datasets are small (tens to a few hundred
cases) and stratified over imbalanced categories; and consultants must
be able to defend every automated decision they sign.

A rich line of work addresses RAG \emph{measurement}. Frameworks such
as RAGAs~\cite{es2024ragas} and ARES~\cite{saadfalcon2024ares}
decompose quality into retrieval-side and generation-side components
scored by an LLM judge~\cite{zheng2023judging,liu2023geval}. Benchmarks
such as RGB~\cite{chen2024benchmarking} and
RAGBench~\cite{friel2024ragbench} assess the evaluators themselves.
Deploying these components inside an accountable gate exposed three
gaps between scoring and deciding. First, \emph{evidence is routinely incomplete}: traces
lack contexts, providers fail, judges return malformed output; averaging
over whatever is available conflates ``measured as bad'' with ``not
measured''. Second, \emph{point estimates overstate certainty} on the
small, stratified evaluation sets that real engagements afford. Third,
\emph{the measuring instrument is unvalidated}: LLM judges exhibit
biases and task-dependent reliability~\cite{wang2024fair,bavaresco2024llms},
so agreement measured on one domain does not transfer to another.

The contributions of this work are fourfold and transferable to any
organisation operating RAG gates.

First, evidence-first decision semantics deployed in production: a
four-state decision lattice (\spromote{}, \sreview{}, \sblock{},
\snoteval{}) with per-metric decision modes and explicit treatment of
missing evidence and judge protocol errors
(Sect.~\ref{sec:decision}). Second, a stratified beta-binomial gate:
per-stratum credible intervals and a probabilistic regression risk
$P(p_B < p_A - \delta)$ replace point estimates
(Sect.~\ref{sec:bayes}). Third, per-engagement judge validation: a
meta-evaluation protocol with balanced golden sets and threshold-free
agreement statistics, mandatory before judge scores may influence
decisions (Sect.~\ref{sec:metaeval}). Fourth, a public evaluation at
two system boundaries: judge-layer validation on RAGBench
(Sect.~\ref{sec:ragbench}), operating characteristics of the
production statistical gate under regression, no change, and
improvement (Sect.~\ref{sec:gate-operating-characteristics}).

Each component builds on established work. The contribution of
\sysname{} is the decision layer that makes missing evidence, judge
protocol errors, statistical uncertainty, and per-engagement judge
validity explicit inputs to a release decision, not hidden assumptions
behind a score.

The paper is organized as follows. Sect.~\ref{sec:related} reviews
related work on RAG evaluation and LLM-as-a-judge reliability.
Sect.~\ref{sec:method} presents the \sysname{} framework: evidence
model, layered scoring, decision semantics, statistical gate, and
judge meta-evaluation. Sect.~\ref{sec:deployment} describes the
deployed platform and the engagement workflow. Sect.~\ref{sec:exp}
reports the evaluation: judge-layer validation on RAGBench and
operating characteristics of the statistical gate.
Sect.~\ref{sec:concl} concludes.

\section{Related Work}\label{sec:related}

\paragraph{RAG evaluation frameworks.}
RAGAs~\cite{es2024ragas} introduced reference-free, LLM-scored metrics
for RAG (faithfulness, answer relevance, context precision and recall).
ARES~\cite{saadfalcon2024ares} trains lightweight judges on synthetic
data and calibrates them with prediction-powered inference over a
small human-labelled set. It targets evaluator accuracy on benchmarks;
\sysname{} instead embeds evaluator validation inside an end-to-end
decision pipeline with explicit missing-evidence semantics. Open-source tooling
popularised the RAG triad of context relevance, adherence, and
answer relevance, which we adopt as the judge-facing decomposition.
RAGAs, ARES, and TruLens provide scores or evaluator predictions rather
than the evidence-aware release semantics studied here; direct
comparison of release decisions would therefore require imposing a
common decision policy.
Toolkits such as NeMo Guardrails~\cite{rebedea2023nemo} provide
programmable input and output rails; \sysname{} incorporates comparable
local guardrails as one metric provider among several, subject to the
same decision semantics.

\paragraph{Benchmarks for RAG and its evaluators.}
RGB~\cite{chen2024benchmarking} probes LLM behaviour under noisy or
counterfactual retrieval. RAGBench~\cite{friel2024ragbench} releases
100k (question, documents, response) traces from 12 datasets across
five industry domains, annotated with the TRACe schema: continuous
\emph{relevance}, \emph{utilization}, and \emph{completeness} scores
and a binary \emph{adherence} label. The labels are produced by LLM
annotators calibrated against human annotations; we therefore refer to
them as \emph{reference labels} rather than ground truth. RAGBench
frames evaluator quality as a supervised prediction problem (RMSE for
continuous targets, AUROC for adherence), letting heterogeneous
evaluators be compared on equal footing; we adopt this protocol.

\paragraph{LLM-as-a-judge reliability.}
LLM judges correlate with human preferences on some tasks
\cite{zheng2023judging,liu2023geval} but exhibit position and verbosity
biases~\cite{wang2024fair}, and large-scale studies report agreement
varying substantially across tasks and domains~\cite{bavaresco2024llms}.
This non-transferability motivates our per-engagement validation:
rather than certifying a judge model once, \sysname{} re-measures
judge--human agreement for every deployment domain, using classical
agreement statistics~\cite{cohen1960coefficient,landis1977measurement},
imbalance-robust summaries~\cite{chicco2020advantages}, and bootstrap
uncertainty~\cite{efron1994introduction}. Existing tooling scores
answers. Converting scores into accountable release decisions under
incomplete evidence is the gap deployed systems face, and the gap this
work addresses.

\section{The \sysnamefull{} Framework}\label{sec:method}

\subsection{Setting and Evidence Model}\label{sec:setting}

An \emph{evaluation case} $x = (q, E, K^\ast, T, S, P)$ consists of a
question $q$, an optional expected behaviour $E$ (reference answer,
expected terms $T$, expected sources $S$), optional retrieval
requirements $K^\ast$, and forbidden phrases $P$. A versioned dataset
$D = \{x_i\}_{i=1}^{n}$ is assembled from observed production traces;
each case carries a category $c(x)$ and a criticality flag. Invoking
the system under test on $x_i$ yields an answer $a_i$ and retrieved
contexts $K_i$ (possibly empty).

A set of \emph{metric providers} maps $(x_i, a_i, K_i)$ to metric
results: each result $m$ carries a name, an optional numeric value $v$,
an optional threshold $\theta$, a boolean pass flag, and the emitting
provider. All thresholds, synonym policies, and judge instructions are
versioned in an \emph{evaluation profile}, never hard-coded: every
decision is reproducible from (dataset snapshot, profile snapshot, code
version). The framework distinguishes four evidence types
throughout: \emph{observed}, \emph{declared}, \emph{inferred} (flagged
as such), and \emph{missing}. Missing evidence must surface as an
explicit gap with an impact statement, never as a default value.

\subsection{Layered Scoring}\label{sec:scoring}

\paragraph{Deterministic checks.}
The first layer is local, cheap, and exactly reproducible: answer
presence, presence of required retrieved contexts, matching of
expected sources and document versions, detection of forbidden
phrases, and matching of expected terms with normalisation of Unicode,
case, and accents plus profile-versioned synonym groups. Term matching
passes when the matched fraction reaches a profile-defined ratio.
These checks provide a judge-independent baseline and catch structural
failures early.

\paragraph{Local guardrails.}
A rule engine inspired by production rail systems~\cite{rebedea2023nemo}
evaluates pre- and post-conditions (PII, secrets, prompt injection,
forbidden phrases, JSON and link validity) with per-rule actions
$\{\textsf{block}, \textsf{warn}, \textsf{log}\}$, timeouts, and
fail-open/fail-closed behaviour. Every rule execution emits a metric
result, so a guardrail timeout is visible evidence rather than a silent
skip.

\paragraph{LLM-judge RAG triad.}
The third layer scores each case on the RAG triad of context relevance
$c$, adherence $g$, and answer relevance $r$ through a single
constrained JSON call to an exchangeable, OpenAI-compatible judge
endpoint at temperature $0$. A metric passes if its score meets the
profile threshold $\theta$. Two protocol rules are central. First,
scores are accepted only within a small tolerance of the declared
range: values in $[-0.01, 1.01]$ are clamped to $[0,1]$; non-numeric,
non-finite, or genuinely out-of-range values are \emph{provider
protocol errors}. The metric is then marked failed with an explanatory
comment and the case is routed to \sreview{}. Malformed judge output
is never coerced to a valid score. A silent zero is indistinguishable
from a measured catastrophic failure, and a silent clamp could convert
an invalid output into a confident pass. Second, adherence is
scored only for cases that require retrieval; for purely
conversational turns the dimension is undefined and omitted rather
than imputed.

At run level, the primary grounding measure is the \emph{adherence
rate}: the fraction of evaluated cases that passed the adherence
criterion.
Stakeholders thus receive an interpretable proportion, not an average
of uncalibrated judge scores; the raw scores remain per-case
diagnostics.

\subsection{Decision Semantics}\label{sec:decision}

Let $M(x)$ be the metric results for case $x$ and
$F(x) = \{ m \in M(x) : \neg m.\mathrm{passed} \}$ the failed subset.
The profile assigns each metric a decision mode
$\mu(m) \in \{\textsf{observe}, \textsf{review}, \textsf{block}\}$;
\textsf{observe} metrics are reported but never influence decisions. A
fixed, documented set $C$ of \emph{critical metric identifiers}
(missing required contexts, failed source or version matching,
forbidden phrases, PII, secret, or injection findings, output-format
violations) escalates independently of mode. The item decision is
\begin{equation}
\mathrm{dec}(x) =
\begin{cases}
\snoteval & M(x) = \emptyset,\\[2pt]
\sblock & \exists\, m \in F(x): \mu(m) = \textsf{block} \;\lor\; \mathrm{name}(m) \in C,\\[2pt]
\sreview & F(x) \setminus \{ m : \mu(m) = \textsf{observe} \} \neq \emptyset,\\[2pt]
\spromote & \text{otherwise,}
\end{cases}
\label{eq:itemdecision}
\end{equation}
and run-level aggregation is conservative:
\begin{equation}
\mathrm{dec}(D) =
\begin{cases}
\sblock & \exists\, i : \mathrm{dec}(x_i) = \sblock,\\[2pt]
\snoteval & \forall\, i : \mathrm{dec}(x_i) = \snoteval,\\[2pt]
\sreview & \exists\, i : \mathrm{dec}(x_i) \in \{\sreview, \snoteval\},\\[2pt]
\spromote & \text{otherwise.}
\end{cases}
\label{eq:rundecision}
\end{equation}
The set $C$ is a fixed system safety floor: membership in $C$ takes
precedence over a per-metric profile mode and therefore blocks even if
that metric was mistakenly configured as observe-only. Outside $C$, the
versioned evaluation profile determines whether a failed metric is
blocking, review-only, or observational; this precedence is part of the
decision semantics, not a profile-dependent risk preference. In
deployment, profile-level blocking is reserved for critical violations
(e.g.\ PII leakage), and every non-\spromote{} outcome carries
machine-readable failure reasons and is routed to consultant review.
Two properties follow. \emph{No implicit pass}: absent evidence can only produce
\snoteval{} or \sreview{}, never \spromote{}. \emph{Human-in-the-loop
by construction}: the gate is an escalation mechanism, not an
autonomous verdict.

\subsection{The Stratified Hierarchical Beta-Binomial
Gate}\label{sec:bayes}

\sysname{} models release quality at the operational levels the gate
acts on. Item decisions are grouped into engagement-defined strata.
Each stratum receives a beta-binomial posterior. Run-level quality is
a weighted posterior aggregate over strata, and baseline and candidate
runs are compared at the posterior level. The hierarchy is
operational---item $\to$ stratum $\to$ run $\to$ comparison---not a
global partial-pooling Bayesian model. The implementation is
deliberately simple, auditable, and profile-configurable.

Point pass-rates on small, stratified evaluation sets are misleading.
Per category $c$, we model the item outcome
$X_i = \mathbf{1} [\mathrm{dec}(x_i) = \spromote]$ as Bernoulli with
rate $p_c$ under a conjugate prior
$p_c \sim \mathrm{Beta}(\alpha_0, \beta_0)$ (default
$\mathrm{Beta}(1,1)$)~\cite{gelman2013bayesian}; observing $k_c$
successes among $n_c$ cases gives the posterior
$\mathrm{Beta}(\alpha_0 + k_c,\, \beta_0 + n_c - k_c)$, reported with
equal-tailed $95\%$ credible intervals per stratum. The run summary is
a weighted posterior mean $\bar{p} = \sum_c w_c\, \mathbb{E}[p_c \mid
\mathrm{data}]$ with weights defaulting to dataset coverage
$w_c = n_c/N$. Since evaluation sets are often deliberately enriched
with critical cases, coverage weights reflect the \emph{evaluation
design}, not production traffic; profiles may therefore supply explicit
target weights (traffic shares or business criticality) when a
production-facing estimate is required, and the report states which
weighting is in effect.

For version comparison, let $A$ (baseline) and $B$ (candidate) be runs
on the same dataset and profile. Drawing $S$ Monte Carlo samples of the
stratified weighted rate for each run (fixed, recorded seeds), the
regression risk is
\begin{equation}
  \rho = P\bigl(p_B < p_A - \delta \mid \mathrm{data}\bigr),
\label{eq:regression}
\end{equation}
with $\delta$ the minimal regression of practical relevance (default
$0.03$); the statistical gate returns \sblock{} if $\rho \ge \tau_b$,
\sreview{} if $\rho \ge \tau_r$ (defaults $0.8$, $0.5$), and
\spromote{} otherwise. It supports the operational decision of
Eq.~\eqref{eq:rundecision} and never overrides blocking evidence. The
two runs are modelled as independent. When they share items, positive
correlation means independence \emph{overestimates} the variance of
the difference, so the gate errs toward \sreview{} and \sblock{}. This
approximation is conservative by design: an unnecessary \sreview{}
costs consultant time, while a silently promoted regression costs a
production incident. This conservative treatment is intentional: when
baseline and candidate evidence is too similar, the gate should expose
the uncertainty and route the comparison to human review rather than
silently promote a risky release. The uninformative default
prior is also deliberate. It confines the evidence about a candidate
to the candidate's own run, rather than anchoring it to earlier
versions, and keeps the audit trail free of hidden information.
Engagements with defensible domain history may supply an informative
prior through the versioned profile ($\alpha_0, \beta_0$ are profile
parameters). Likewise, $\delta$ is an engagement-level risk parameter,
not a constant; stratum-specific $\delta$ is a natural profile
extension. All hyperparameters live in the versioned profile and are
reported with each run. Changing the release risk appetite means
changing versioned thresholds, not the evidence model or hidden code
constants.
Sect.~\ref{sec:exp} evaluates judge behavior under the frozen
meta-evaluation protocol and the statistical gate's operating
characteristics under three controlled data-generating scenarios.

\subsection{Per-Engagement Judge Meta-Evaluation}\label{sec:metaeval}

Judge--human agreement is task- and domain-dependent
\cite{bavaresco2024llms,wang2024fair}; \sysname{} therefore treats it
as a quantity to re-measure per engagement. Before judge-based metrics
may influence gate decisions for a new deployment domain, the following
protocol must be executed and its report attached to the engagement
record.

\emph{Golden set.} A domain golden set with binary human labels per
judge dimension: at least 30 cases as an absolute floor for a
smoke-level check, and 100 or more for decision-grade validation; both
classes present for every dimension (minority class $\ge 30\%$); a
deliberate share of borderline cases; independent labelling by two
annotators with adjudication, and inter-annotator agreement reported as
the ceiling against which judge agreement is read.

\emph{Statistics.} Judge scores are binarised at the operating
threshold $\theta$ and compared to labels: raw agreement; Cohen's
$\kappa$~\cite{cohen1960coefficient,landis1977measurement}, reported as
\emph{undefined} when expected agreement is $1$ (single-class labels)
rather than coerced; the Matthews correlation coefficient, informative
under class imbalance~\cite{chicco2020advantages}; threshold-free AUROC
over raw scores; a threshold sweep, since $\theta$ is itself a profile
parameter; and $95\%$ bootstrap confidence
intervals~\cite{efron1994introduction} throughout. A judge-dependent
metric may run in \textsf{block} mode only when its validation report
meets the profile's minimum lower-bound and maximum interval-width
criteria. In practice an engagement starts with
judge metrics in review-only mode and graduates them as the golden set
matures. Validation evidence is bound to the exact judge configuration
(model identifier, frozen rubric, operating threshold). Hosted judge
models change without notice, so switching judge endpoint or model
version invalidates the report and re-triggers the protocol.

\section{Deployment}\label{sec:deployment}

\sysname{} is implemented as a self-hosted platform (API backend and
web console) deployed at an IT consulting firm and in active use for
RAG assessment engagements with enterprise clients. Deployment
constraints shaped the architecture. Client data must not leave the
premises: all storage, scoring, and reporting run inside the client or
consultant perimeter. The only outbound call is the judge endpoint,
exchangeable between a cloud API and a self-hosted OpenAI-compatible
runtime for deployments that forbid egress. Observed traces arrive
from heterogeneous observability stacks and are normalised into a
provider-agnostic schema. Personally identifiable information and
secrets are detected and redacted before a trace can become a dataset
case. Evaluation datasets are curated before any gate run: duplicates,
weak cases, coverage gaps, and unresolved privacy findings are
surfaced, and blocking findings prevent the gate from running at all.

An engagement follows a repeatable workflow: import observed
conversations; curate a versioned dataset; select or adapt an
evaluation profile (thresholds, guardrail policy, judge policy,
statistical policy---all versioned); execute the per-engagement judge
meta-evaluation of Sect.~\ref{sec:metaeval}; run the gate; and deliver
a report. Every decision in the report carries its failure reasons,
evidence types, per-stratum uncertainty, and gaps. Every run persists
content-hashed snapshots of dataset and profile, so a decision remains
auditable after either evolves. Non-promoted cases are reviewed by
consultants; the gate escalates, humans decide.

\section{Evaluation}\label{sec:exp}

We evaluate two claims at their appropriate boundaries. First,
RAGBench tests the judge layer and the frozen meta-evaluation protocol;
it is not an end-to-end validation of \sysname{}. Second, controlled
simulation measures the statistical gate's unsafe-promotion risk,
false-alarm cost, and improvement throughput. All gate-level experiments
are local and make no external model or benchmark calls.

\subsection{Judge-Layer Validation on RAGBench}\label{sec:ragbench}

Engagement data is proprietary, so we evaluate the judge layer on a
public benchmark. RAGBench provides complete observed traces, so it
plugs directly into \sysname{}'s evaluation contract without running
any retrieval system. Each instance becomes a case with question,
answer, and retrieved contexts. We ask:
\emph{how well do the framework's judge metrics agree with the
benchmark's reference labels, how does this depend on the judge model,
and what does it imply for gate operating points?}

\subsubsection{Benchmark and Metric Mapping.}

The benchmark~\cite{friel2024ragbench} is organised into 12 component
datasets across five industry domains (listed in
Table~\ref{tab:perdataset}), with TRACe reference labels produced by
LLM annotators calibrated against human annotations. We map our
adherence score to the binary \texttt{adherence} label, compared by
AUROC and by MCC/F1 at the deployed operating threshold
$\theta = 0.8$. Our context-relevance score is holistic, whereas the
TRACe \texttt{relevance} label is a fraction of relevant context spans;
the definitions differ, so rank correlations (Spearman, Kendall) are
the primary comparison and RMSE is reported only as supplementary
material with this caveat. Answer relevance has no TRACe counterpart
and is excluded; utilization and completeness are not natively produced
by our judge and are left to future work.

\subsubsection{Protocol.}

All reported numbers use stratified samples of $100$ instances per
component dataset, balanced by adherence label where possible, with
fixed published seeds. These samples support controlled comparison,
not population-level performance estimates. Sample manifests (exact instance identifiers)
are persisted, so every judge scores identical instances. Judge
responses are cached by (instance, model, rubric hash). The rubric was
frozen before any test-split instance was scored; no prompt or
threshold was adjusted afterwards. Judge model selection is part of
the protocol: candidate judges (\texttt{gpt-4.1-nano},
\texttt{gpt-4o}; temperature $0$) were compared on a
\emph{validation}-split sample of three datasets ($N{=}300$), with MCC
at the deployed threshold as the pre-specified primary criterion. The
selected judge and the low-cost judge were then run once on the
\emph{test} split of all 12 datasets ($N{=}1200$ per judge). All
statistics carry $95\%$ bootstrap confidence intervals ($1{,}000$
resamples, fixed seed). Judge protocol errors are excluded from
agreement statistics and reported separately; none occurred in the
reported runs.

Pre-registered sanity audits caught two pipeline defects before the
reported runs: a payload bug silently capping the number of documents
passed to the judge, and duplicate instance identifiers across
component datasets. The latter forced aborting a first full-test
attempt and discarding its partial results. Payload audits confirm
that in all reported runs every document of every instance reached the
judge.

\subsubsection{Results.}

\begin{table}[t]
\centering
\caption{Judge selection on the validation split (3 datasets,
$N{=}300$ per judge; $95\%$ bootstrap CIs). MCC is computed at the
deployed threshold $\theta{=}0.8$.}
\label{tab:selection}
\footnotesize
\setlength{\tabcolsep}{3pt}
\begin{tabular}{lcccc}
\toprule
Judge & AUROC adh. & MCC @ $0.8$ & Spearman rel. & Cost \\
\midrule
gpt-4.1-nano & $0.672$ \ci{0.609}{0.732} & $0.284$ \ci{0.176}{0.391} & $0.279$ \ci{0.163}{0.386} & \$0.06 \\
gpt-4o       & $0.787$ \ci{0.731}{0.837} & $0.435$ \ci{0.329}{0.539} & $0.217$ \ci{0.103}{0.330} & \$1.36 \\
\bottomrule
\end{tabular}
\end{table}

\paragraph{Judge selection.}
Table~\ref{tab:selection} reports the validation comparison. By the
pre-specified criterion (MCC at the operating threshold), gpt-4o was
selected for the full test run. The same protocol makes judge
cost-quality trade-offs directly measurable per engagement.

\begin{table}[t]
\centering
\caption{Test-split results over all 12 RAGBench datasets ($N{=}1200$
per judge, identical instances; $95\%$ bootstrap CIs; overall micro
aggregation).}
\label{tab:test}
\footnotesize
\setlength{\tabcolsep}{3pt}
\begin{tabular}{lcccc}
\toprule
Judge & AUROC adh. & MCC @ $0.8$ & Spearman rel. & Cost \\
\midrule
gpt-4.1-nano & $0.603$ \ci{0.570}{0.634} & $0.170$ \ci{0.110}{0.221} & $0.366$ \ci{0.313}{0.413} & \$0.39 \\
gpt-4o       & $0.783$ \ci{0.756}{0.807} & $0.433$ \ci{0.386}{0.484} & $0.298$ \ci{0.243}{0.351} & \$9.42 \\
\bottomrule
\end{tabular}
\end{table}

\paragraph{Judge quality dominates.}
Table~\ref{tab:test} shows the headline result. On identical test
instances, gpt-4o exceeds the low-cost judge by $+0.180$ AUROC and
$+0.263$ MCC at the deployed threshold, with non-overlapping confidence
intervals. The low-cost judge detects non-adherent answers only weakly
above chance (AUROC $0.603$). Its per-class score distributions
overlap heavily; no threshold in a $0.1$--$0.9$ sweep yields MCC above
$0.17$, so the deficiency is not a threshold artifact. Nothing in its
operational output signals this: across all runs it produced zero
protocol errors and well-formed, plausible explanations. An earlier
pilot on three datasets ($N{=}300$) had yielded AUROC $0.563$
\ci{0.495}{0.631} for the same judge. That interval includes chance,
which is why the framework mandates confidence intervals before
conclusions are drawn.
\begin{figure}[t]
\centering
\includegraphics[width=\textwidth]{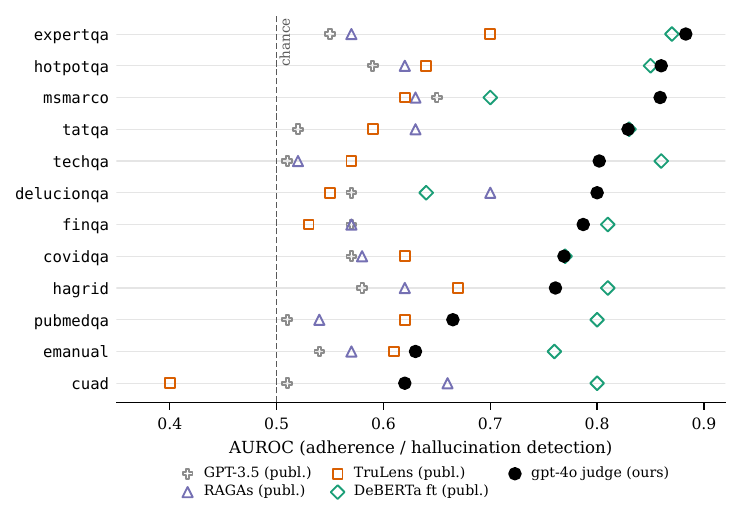}
\caption{Per-dataset adherence AUROC of the selected judge (gpt-4o,
filled circles; stratified test samples, $N{=}100$ per dataset)
alongside evaluator results published by the benchmark authors on full
test splits (their Table~3~\cite{friel2024ragbench}). Dashed line: chance.
Datasets sorted by gpt-4o AUROC; the comparison across protocols is
indicative (Sect.~\ref{sec:ragbench}). Per-dataset MCC and Spearman
values appear in Table~\ref{tab:perdataset}.}
\label{fig:perdataset}
\end{figure}
\begin{table}[t]
\centering
\caption{Per-dataset test results for the selected judge (gpt-4o,
$N{=}100$ per dataset).}
\label{tab:perdataset}
\setlength{\tabcolsep}{5pt}
\begin{tabular}{lccc}
\toprule
Dataset & AUROC (adherence) & MCC @ $0.8$ & Spearman (relevance) \\
\midrule
\texttt{covidqa}    & $0.769$ & $0.388$ & $0.009$ \\
\texttt{cuad}       & $0.620$ & $0.199$ & $0.201$ \\
\texttt{delucionqa} & $0.800$ & $0.166$ & $0.236$ \\
\texttt{emanual}    & $0.630$ & $0.040$ & $0.324$ \\
\texttt{expertqa}   & $0.883$ & $0.572$ & $0.400$ \\
\texttt{finqa}      & $0.787$ & $0.446$ & $0.233$ \\
\texttt{hagrid}     & $0.761$ & $0.319$ & $0.128$ \\
\texttt{hotpotqa}   & $0.860$ & $0.530$ & $-0.159$ \\
\texttt{msmarco}    & $0.859$ & $0.618$ & $0.363$ \\
\texttt{pubmedqa}   & $0.665$ & $0.311$ & $0.361$ \\
\texttt{tatqa}      & $0.829$ & $0.563$ & $-0.053$ \\
\texttt{techqa}     & $0.802$ & $0.393$ & $0.583$ \\
\bottomrule
\end{tabular}
\end{table}

\paragraph{Reliability is domain-dependent even for the strong judge.}
Table~\ref{tab:perdataset} disaggregates the selected judge. AUROC
ranges from $0.620$ (\texttt{cuad}, legal contracts) to $0.883$
(\texttt{expertqa}), and at the deployed threshold MCC ranges from
$0.040$ (\texttt{emanual}) to $0.618$ (\texttt{msmarco}): on some
domains the production operating point is close to uninformative even
for a frontier judge. This is the empirical core of the central design
requirement: judge quality cannot be certified once and assumed
elsewhere. It must be measured per engagement, with thresholds
calibrated per domain---exactly what the versioned profiles and the
meta-evaluation protocol implement. The relevance rank correlations
are low and unstable across datasets (including negative values),
consistent with the definitional mismatch noted in the metric mapping
above; we treat this comparison as exploratory.

\paragraph{Positioning against published evaluators.}
The benchmark authors report AUROC for response-level hallucination
detection (equivalently, adherence discrimination) on the full test splits
for a zero-shot GPT-3.5 judge, RAGAs, TruLens, and a DeBERTa-large
encoder fine-tuned on RAGBench (their Table~3~\cite{friel2024ragbench}):
across the 12 component datasets, GPT-3.5 ranges $0.51$--$0.65$ (mean
$0.56$), RAGAs $0.52$--$0.70$ (mean $0.60$), TruLens $0.40$--$0.70$
(mean $0.59$), and the fine-tuned DeBERTa $0.64$--$0.87$ (mean
$0.79$). Under our protocol, the gpt-4o judge (per-dataset mean
$0.77$, micro $0.783$) falls within the range reported for the
fine-tuned encoder and above the zero-shot baselines, while our low-cost judge falls
within the range of the earlier zero-shot evaluators. Published
RAGBench baselines are included for orientation only, not as a
protocol-identical comparison: the published numbers use full test
splits and earlier GPT-3.5-backed evaluators, whereas our judge layer
is evaluated on fixed stratified samples of $100$ instances per
dataset under a frozen judge protocol. The ordering nonetheless
reinforces the central point: evaluator quality varies widely and must
be measured, not assumed. These results support the judge-layer and
meta-evaluation analysis under the tested benchmark conditions; they do
not validate the full release gate.

\subsection{Operating Characteristics of the Statistical
Gate}\label{sec:gate-operating-characteristics}

We exercise the same statistical-benchmark paths used by the product,
\texttt{build\_\allowbreak statistical\_\allowbreak benchmark} and
\texttt{compare\_\allowbreak statistical\_\allowbreak benchmarks}.
Each synthetic run contains $N$ binary item decisions
($N = 20$, $30$, $50$, $100$, or $200$), split across
\texttt{general}, \texttt{domain\_\allowbreak specific}, and
\texttt{critical} strata in proportions $0.50$, $0.30$, and $0.20$.
Their baseline pass probabilities are $0.90$, $0.85$, and $0.80$.
Candidate
probabilities change by $-0.05$, $0$, or $+0.05$ in every stratum,
defining the \emph{regressed}, \emph{unchanged}, and \emph{improved}
scenarios. For each scenario and $N$, we run 1,000 replicates with
data seed 20260709. Baseline and candidate observations are
independent within each replicate. The beta-binomial gate uses a
$\mathrm{Beta}(1,1)$ prior, $95\%$ credibility, minimum relevant
regression $\delta=0.03$, block threshold $0.80$, 121 posterior
samples, and independent posterior streams seeded at 4201 and 4202.
Run-level posterior draws use the empirical stratum fractions induced
by the stated allocation.

We compare a naive gate, which promotes when the observed candidate
pass rate is at least the observed baseline pass rate minus $\delta$,
with two \sysname{} profiles. \texttt{balanced\_default} has review
threshold $0.50$; \texttt{decision\_grade}, used in
Table~\ref{tab:gate-operating-characteristics}, has review threshold
$0.40$. Both profiles receive exactly the same item observations and
posterior regression probability in each replicate, so their only
controlled difference is the review threshold. The seed schedule was
fixed in an earlier regression-only pilot of the same generator and
reused unchanged when the two additional scenarios were added; it is
not the result of a seed search. It does not create a paired Bayesian
model: baseline and candidate item outcomes, and their posterior
draws, remain independent.

In these synthetic scenarios, promotion of a regressed candidate is an
unsafe promotion.
Under no true change, manual review or block is a false alarm. Under
true improvement, promotion measures release throughput. We report all
three outcomes together to expose both the protection and the
operational cost of each policy.

\begin{table}[t]
\centering
\caption{Gate operating characteristics for the
\texttt{decision\_grade} profile (rates in \%).}
\label{tab:gate-operating-characteristics}
\footnotesize
\begin{tabular}{r|r|r|r|r|r}
\toprule
$N$ & Naive promote & AGO promote & AGO review & AGO block & Avg. $\rho$ \\
\midrule
\multicolumn{6}{l}{\textit{Regressed}} \\
20 & 41.8 & 33.0 & 47.3 & 19.7 & 0.532 \\
30 & 37.4 & 35.1 & 41.3 & 23.6 & 0.538 \\
50 & 38.1 & 34.5 & 40.8 & 24.7 & 0.546 \\
100 & 38.4 & 29.9 & 42.5 & 27.6 & 0.571 \\
200 & 29.3 & 22.2 & 41.6 & 36.2 & 0.638 \\
\addlinespace
\multicolumn{6}{l}{\textit{Unchanged}} \\
20 & 59.7 & 51.4 & 38.3 & 10.3 & 0.419 \\
30 & 56.3 & 52.7 & 38.1 & 9.2 & 0.408 \\
50 & 67.9 & 60.7 & 31.7 & 7.6 & 0.356 \\
100 & 76.6 & 70.0 & 25.2 & 4.8 & 0.286 \\
200 & 82.1 & 76.2 & 21.1 & 2.7 & 0.249 \\
\addlinespace
\multicolumn{6}{l}{\textit{Improved}} \\
20 & 76.7 & 71.5 & 24.4 & 4.1 & 0.297 \\
30 & 82.3 & 78.9 & 19.6 & 1.5 & 0.245 \\
50 & 89.7 & 85.1 & 13.5 & 1.4 & 0.176 \\
100 & 97.7 & 95.9 & 4.0 & 0.1 & 0.081 \\
200 & 99.8 & 99.4 & 0.6 & 0.0 & 0.028 \\
\bottomrule
\end{tabular}
\end{table}

Table~\ref{tab:gate-operating-characteristics} shows a consistent
decision-grade reduction in unsafe promotion under true regression.
Across $N=20$--$200$, the naive gate promotes $29.3\%$--$41.8\%$ of
regressed candidates, whereas \sysname{} promotes $22.2\%$--$35.1\%$;
the reduction holds at every tested $N$. The protection is not free.
Under no true change, the decision-grade profile promotes
$51.4\%$--$76.2\%$ and sends $23.8\%$--$48.6\%$ to review or block.
Under true improvement, it promotes $71.5\%$--$99.4\%$, compared with
$76.7\%$--$99.8\%$ for the naive rule. As $N$ grows, the average
posterior regression probability separates the scenarios: at $N=200$
it is $0.638$ for regression, $0.249$ for no change, and $0.028$ for
improvement.

The balanced profile is reported in the accompanying artifact rather
than hidden. Under regression it promotes $29.8\%$--$43.9\%$, which
can match or exceed the naive rate, while under no change it promotes
$60.6\%$--$83.0\%$ with a $17.0\%$--$39.4\%$ false-alarm rate. Under
improvement it promotes $77.4\%$--$99.8\%$. Because the two profiles
share all simulated evidence, the stronger regression protection of
\texttt{decision\_grade} is attributable to its lower review
threshold, not to different samples or a different posterior model.
These operating characteristics describe policy trade-offs under the
tested data-generating process.
\FloatBarrier
\section{Conclusion}\label{sec:concl}

\sysnamefull{} turns RAG evaluation from score production into an
auditable release decision. The RAGBench study shows that judge
reliability must be measured under a frozen meta-evaluation protocol,
not inferred from well-formed output. The three-scenario gate
experiment exposes the release policy's operating characteristics: the
decision-grade profile lowers unsafe promotion at every tested sample
size, at an explicit false-alarm and throughput cost.

Deployment and evaluation yielded four operational findings. First,
judges can fail silently: the low-cost judge produced flawless, plausibly
explained output while discriminating barely above chance; only
measurement against labelled references revealed it. Second,
confidence intervals change conclusions: the $N{=}300$ pilot (AUROC
$0.563$ \ci{0.495}{0.631}) could not be distinguished from chance, and
only the full run supported a verdict. Third, judge reliability is
model- and domain-dependent (MCC $0.040$--$0.618$ at a fixed
threshold); thresholds must be calibrated per engagement, not baked
into code. Fourth, missing evidence must be a first-class outcome;
distinguishing
``measured as bad'' from ``not measured'' is what made the failures
above visible.

These findings support the central claim: evidence completeness, judge
validity, posterior regression risk, and fixed critical precedence
must be first-class release criteria. Future work covers a paired
Bayesian comparison, native evaluation of the remaining TRACe
dimensions, and repeated-sampling analysis of per-case variance.

\subsubsection*{Use of Generative AI.}
Generative AI tools were used to assist with drafting and editing the
manuscript and with implementing the evaluation harness and simulation
code. All content, code, and reported numbers were reviewed and
verified by the authors, who take full responsibility for them.

\subsubsection*{Ethical Considerations.}
The judge-layer study uses the public RAGBench benchmark. The
statistical operating-characteristic study uses synthetic binary
outcomes. No personal, client, or proprietary data are introduced by
either gate-level experiment, and neither experiment
calls an external model service. The deployed framework targets
enterprise settings and includes local redaction of personally
identifiable information before traces are persisted; gate outcomes are
decision support with mandatory human review of non-promoted cases, not
autonomous verdicts.

\bibliographystyle{splncs04}
\bibliography{riferimenti}

\end{document}